\documentclass{article}

\PassOptionsToPackage{numbers,compress}{natbib}
\usepackage[preprint,dblblindworkshop]{neurips_2026}
\workshoptitle{Can We Trust the Judge?}

\usepackage[utf8]{inputenc}
\usepackage[T1]{fontenc}
\usepackage{hyperref}
\usepackage{url}
\usepackage{booktabs}
\usepackage{amsmath}
\usepackage{amsfonts}
\usepackage{graphicx}
\usepackage{xcolor}
\usepackage{microtype}

\newcommand{\matchthreshold}{0.70}

\title{More Than Mimicking Reviewers:\\
Evaluating LLMs for Pre-Submission Peer Review}

\author{
\begin{tabular}[t]{c}
\textbf{Pouya Parsa}\thanks{Equal contribution.} \\[2pt]
{\small University of Minnesota} \\
{\small\texttt{parsa025@umn.edu}}
\end{tabular}
\And
\begin{tabular}[t]{c}
\textbf{Amin Rezaei}\footnotemark[1] \\[2pt]
{\small Invariant Tech Inc.} \\
{\small\texttt{amin@invariant.sh}}
\end{tabular}
}

\begin{document}

\maketitle

\begin{abstract}
Peer-review feedback often arrives too late for authors to make meaningful revisions. We study an
author-facing LLM system that moves part of this stress test before submission:
it generates a broad pool of atomic concerns and compresses
them into a short report. We evaluate agreement with historical reviews and,
separately, the possible validity of concerns they omit.

From 10,000 ICLR 2026 submissions, we use 3,398 manuscripts with accessible
versions that predate review. On a ten-paper diagnostic, independent sampling
covers 44.9\% of historical issues; deduplication and refill reaches 78.7\%
strict and 84.9\% seriousness-weighted coverage, at 3.6$\times$ more requests
and 5.2$\times$ more tokens. A hidden Top-32 Oracle preserves the full 79.3\%
weighted coverage of a 256-candidate pool, but paper-only selectors retain only
40--44\%. LLM review therefore provides broad coverage with a large candidate
pool but compresses poorly; ablations identify representative selection and
matcher sensitivity as the main sources of this gap.
\end{abstract}

\section{Introduction}

Formal peer review supplies some of the most valuable feedback a research idea
will receive, but usually only after submission. By then, there may be little
time to add a missing experiment, repair a central claim, or redesign an
evaluation. We ask whether an author-facing LLM system can move part of this
stress test earlier: inspect a draft, anticipate concerns plausible reviewers
may raise, and surface additional weaknesses while revision is still possible.

Our system follows a broad-then-compact design. It first generates hundreds of
evidence-grounded concerns, removes semantic repeats, and then selects a short
report. This design couples generation to judging: the generator bounds what
can be found, while semantic decisions determine which concerns count as new,
which survive compression, and which pipeline appears best. Known position and
self-preference effects \citep{wang2023fair,panickssery2024selfrecognition}
therefore become system risks, not merely evaluation artifacts.

Historical reviews also form an incomplete reference. A small panel samples
only some valid criticisms. A generated issue that matches a historical review
demonstrates \emph{reviewer anticipation}; an unmatched issue may be invalid,
or it may be useful feedback the panel missed. We consequently separate
\emph{historical-review agreement} from \emph{independent validity}. This
distinction is central to the intended product: broad AI feedback should
overlap strongly with human feedback without being forced to imitate it
exactly.

We evaluate on reviewer-visible ICLR 2026 manuscripts. Paper-only generation is
frozen before human reviews are opened; reviews are then converted to corrected
atomic issues, and every candidate is compared with every issue. This protocol
measures breadth, redundancy, historical coverage, compression loss, and judge
disagreement without revealing review targets during generation. The results
show that broad generation can recover much of a finite panel's issue weight,
but reducing the pool to 32 useful concerns remains difficult.

Our contributions are:
\begin{itemize}
  \item a leakage-controlled benchmark built from 10,000 submissions and a
        strict cohort of 3,398 pre-review manuscripts;
  \item a controlled pipeline comparison showing that deduplication and
        refill raise weighted historical coverage from 49.5\% to 84.9\%, but
        cost 5.2$\times$ more tokens;
  \item a Top-32 study that separates three error sources: semantic
        clustering, representative selection, and matching/atomization; and
  \item evidence that the AI and human concern sets overlap substantially but
        are not interchangeable, motivating separate agreement and validity
        evaluation.
\end{itemize}

\section{Related work}

\paragraph{Review data and simulation.}
PeerRead enabled data-driven study of 14.7K drafts and 10.7K expert reviews
\citep{kang2018peerread}; later work uses LLM agents to simulate reviewer
behavior and decisions \citep{jin2024agentreview}. We instead study an
author-facing task: discovering atomic concerns from a draft and retaining a
compact subset.

\paragraph{LLM-generated scientific feedback.}
GPT-4 feedback can overlap with human comments at rates comparable to
human--human overlap, and many authors report finding it useful
\citep{liang2024feedback}. MARG uses specialist agents
\citep{darcy2024marg}, OpenReviewer fine-tunes an 8B reviewer on 79K reviews
\citep{idahl2025openreviewer}, and TreeReview expands review questions
dynamically \citep{chang2025treereview}. Our focus is complementary: semantic
breadth, generation cost, Top-$k$ compression, and atomic agreement with hidden
human issues.

\paragraph{LLM-as-a-judge.}
LLM judges capture semantic equivalence better than lexical metrics
\citep{zheng2023judging,liu2023geval}, but exhibit position and same-family
biases \citep{wang2023fair,panickssery2024selfrecognition}. We study these
failures when judge scores define both evaluation and selection.

\section{Methods}
\label{sec:methods}

\subsection{Task and system design}

A draft goes through two main stages in our system: broad candidate generation
and compression into an author-facing report.

\paragraph{Broad generation.}
For manuscript $P$, a paper-only generator produces
$C(P)=\{c_1,\ldots,c_K\}$. Independent sampling is the simplest baseline, but
it repeatedly finds common concerns. Our deduplication-and-refill pipeline
checks schema and paper evidence, rejects a candidate judged semantically near
an accepted concern, supplies the generator with a compact exclusion list, and
continues until the candidate quota is filled. This favors marginal semantic
breadth rather than the number of nominal outputs.

\paragraph{Compact selection.}
The paper-only selector $S_k(P,C)\subseteq C(P)$ must choose at most $k=32$
concerns without access to historical reviews. We compare direct paper-only
ranking with cluster-then-select methods: partition candidates by semantic
similarity, score possible representatives using only the manuscript, and
allocate the final slots across clusters. We call selectors that use hidden
review matches \emph{Oracles}: they provide upper bounds but are not feasible at
inference time.

\subsection{Historical-review agreement}

Historical reviews are transformed into deduplicated issues
$H(P)=\{h_1,\ldots,h_J\}$. For candidates $c_i$, a semantic judge estimates
same-concern probability $M_{ij}$, where equivalence means the same substantive
concern rather than a shared category. We measure issue recall $R(C,H)$ at
threshold $\tau=\matchthreshold$:

\begin{equation}
  \small
  M_{ij}=\Pr(c_i\equiv h_j\mid P,c_i,h_j),\qquad
  R(C,H)=\frac{1}{|H|}\sum_{j=1}^{|H|}
  \mathbf{1}\!\left[\max_i M_{ij}\ge\tau\right].
\end{equation}

Weighted coverage replaces issue counts with seriousness weights $w_j$; we also
report Major-Issue Recall, candidate match rate, and semantic uniqueness. Hit
rate differs from coverage because repeated candidates can match one issue
while leaving others undiscovered. The unrestricted Top-$k$ Oracle maximizes
covered weight using $M$; the cluster Oracle first fixes a partition. Their gap
measures partition loss, and the remaining gap to a paper-only selector measures
representative-selection loss.

\subsection{Independent validity and semantic saturation}

The match matrix measures \emph{reviewer anticipation}, not validity. Unmatched
concerns require separate assessment for grounding, soundness, specificity,
importance, and whether the paper already addresses them. Semantic saturation
occurs as novel concerns become harder to find: if $p_n$ is the acceptance
probability after $n$ concerns, reaching $K$ requires
$\mathbb{E}[A_K]=\sum_{n=0}^{K-1}1/p_n$ attempts in expectation. Generator
mixtures help only when models cover different semantic regions.

\section{Experiments}
\label{sec:experiments}

\subsection{Dataset and manuscript recovery}

\paragraph{OpenReview source.}
We ingest a read-only snapshot of the public ICLR 2026 record from OpenReview
API v2.\footnote{\url{https://api2.openreview.net}; conference group:
\url{https://openreview.net/group?id=ICLR.cc/2026/Conference}.} The snapshot
contains 10,000 submissions and 37,993 official reviews;
9,999 current forum PDFs pass signature validation.

\paragraph{Reviewer-visible version recovery.}
The current forum PDF can contain post-review changes, so it is not used as the
input. OpenReview--arXiv mapping recovers 2,123 pre-submission versions, while
OpenReview timestamps identify 1,548 pre-review PDFs through a second recovery
path. The paths overlap for 273 papers; selecting arXiv on overlap yields 3,398
reviewer-visible manuscripts.

\paragraph{Post-review and publication-side snapshot.}
We retain each paper's current OpenReview PDF separately from its pre-review
input. For 1,392 accepted papers, this provides a public post-review version for
future revision analysis.

\begin{figure}[t]
  \centering
  \includegraphics[width=\linewidth]{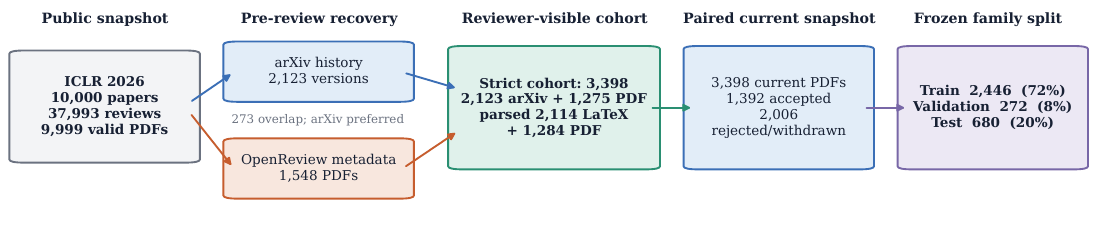}
  \caption{Data construction pipeline.}
  \label{fig:data-collection}
\end{figure}

Figure~\ref{fig:data-collection} summarizes the recovery and version-pairing
workflow. Historical and current files remain separate throughout.

\begin{table}[t]
  \caption{Corpus construction and frozen paper-family split. Decisions are
  normalized from the explicit decision note and current venue state.}
  \label{tab:dataset}
  \centering
  \small
  \setlength{\tabcolsep}{4.2pt}
  \begin{tabular}{@{}lrrrrr@{}}
    \toprule
    Cohort & Oral & Poster & Reject & Withdrawn & Total \\
    \midrule
    OpenReview snapshot & 224 & 5,127 & 4,473 & 176 & 10,000 \\
    Strict historical cohort & 61 & 1,331 & 1,940 & 66 & 3,398 \\
    \midrule
    Train & 44 & 958 & 1,397 & 47 & 2,446 \\
    Validation & 5 & 106 & 155 & 6 & 272 \\
    Held-out test & 12 & 267 & 388 & 13 & 680 \\
    \bottomrule
  \end{tabular}
\end{table}

\paragraph{Frozen split.}
Duplicate and likely resubmitted papers share a family ID before stratification.
The immutable split contains 2,446 training, 272 validation, and 680 held-out
test papers (Table~\ref{tab:dataset}). This paper reports retrospective
training diagnostics and one five-paper validation pilot; test targets remain
unopened.

\subsection{Evaluation protocol}

\paragraph{Atomic targets.}
Reviews are split into manually corrected, deduplicated atomic concerns.
Seriousness distinguishes quick fixes from issues requiring new experiments or
changes to central claims. The ten-paper diagnostic contains 178 issues.

\paragraph{Leakage boundary.}
Generation uses only the historical manuscript or its frozen structured
representation; reviews, ratings, rebuttals, decisions, later versions, and
match matrices remain hidden.

\paragraph{Generators and judges.}
Generators and selectors include GPT-5.6 Sol, DeepSeek V4 Flash, Inkling
NVFP4, Kimi-K3, and Qwen3.8-2.4T-A95B. Matching uses schema-constrained,
temperature-zero calls and preserves the complete matrix.

\subsection{Baselines and comparisons}

At 512 candidates, we compare independent sampling with deduplication and
refill, then test early stopping and model mixtures. At 256 candidates, we
compare paper-only compression with Top-32 Oracles; a second matcher tests judge
dependence. Some settings vary jointly, so these are pipeline comparisons.

\subsection{Broad candidate generation}
\label{sec:generation-results}

\subsubsection{Independent sampling versus deduplication and refill}

Table~\ref{tab:strategy-comparison} compares two DeepSeek V4 Flash pipelines at
512 retained candidates per paper. Independent persona-batched calls preserve
repeated answers. Deduplication and refill rejects a concern near an accepted
one, supplies a compact forbidden-concern list, and continues until quotas are
filled.

\begin{table}[t]
  \caption{DeepSeek generation strategies on ten papers at 512 retained
  candidates per paper.}
  \label{tab:strategy-comparison}
  \centering
  \scriptsize
  \setlength{\tabcolsep}{3.8pt}
  \resizebox{\linewidth}{!}{%
  \begin{tabular}{lrrrrrrr}
    \toprule
    Strategy & Covered & Strict & Weighted & Unique & Requests & Tokens & Active req. h \\
    \midrule
    Independent batches & 80/178 & 0.449 & 0.495 & 656/5,120 & 525 & 18.66M & 2.23 \\
    Deduplicate + refill & \textbf{140/178} & \textbf{0.787} & \textbf{0.849} & 5,120 retained & 1,905 & 96.61M & 95.82 \\
    \bottomrule
  \end{tabular}}
\end{table}

Independent sampling covers 80/178 issues. Although 1,265/5,120 candidates hit
at least one target, only 656 (12.8\%) are semantically unique: many hits repeat
the same issues. Deduplication and refill covers 140/178 despite a lower
candidate hit rate (0.204 versus 0.247). The gain is semantic breadth, not more
nominal positives, but it requires 16,143 raw concerns, 1,905 requests, and
96.61M tokens, compared with 525 requests and 18.66M tokens for independent
sampling.

\begin{figure}[t]
  \centering
  \includegraphics[width=\linewidth]{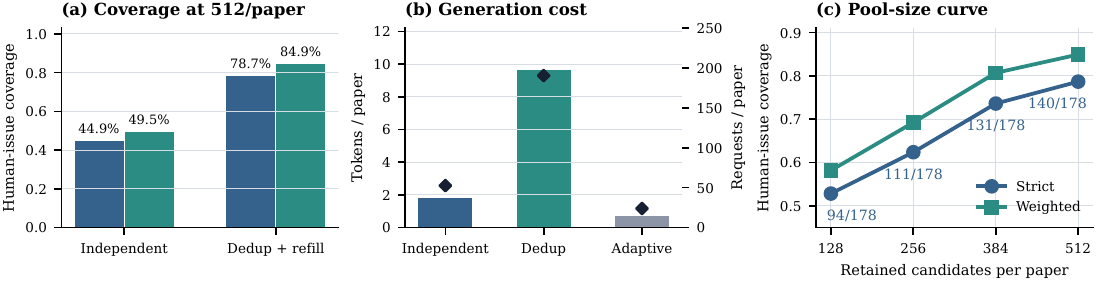}
  \caption{Candidate-generation coverage and cost.}
  \label{fig:generation-tradeoff}
\end{figure}

\subsubsection{Coverage gains become expensive}

As we generate more unique candidates, coverage grows, but requests and token
use rise faster. Expanding from 256 to 512 candidates raises weighted coverage
from .692 to .849 but requires 593 additional requests and 55.61M tokens; the
final 128 candidates add only nine issues. Novelty-based early stopping reduces
requests by 87.6\% and tokens by 92.3\%, but lowers weighted coverage to .660.

\subsubsection{Heterogeneous generators}

On 93 training papers, size-matched 32-candidate Sol and DeepSeek pools obtain
macro strict recall of .387 and .423; their union reaches .535, with 265
Sol-only and 347 DeepSeek-only targets. A provisional 256-candidate mixture
(128 DeepSeek, 64 Inkling, 32 Kimi, 32 Qwen) reaches .767 weighted coverage on
five validation papers. Different policies and budgets prevent attributing the
gain solely to model identity, but the support is complementary.

\subsection{Top-32 compression}
\label{sec:clustering}

The practical report must compress 256 concerns to 32. On ten training papers,
the unrestricted Top-32 Oracle preserves the full pool's .793 weighted
coverage, so 32 slots have sufficient capacity under the frozen matrix.
Paper-only selectors reach only .402--.442. Clustering preserves most of the
coverage available to the unrestricted Oracle; the larger loss occurs when a
paper-only model chooses which member represents each cluster. We isolate the
partition, selection, and matcher effects in Section~\ref{sec:ablations}
(Table~\ref{tab:clustering-selection}).

\begin{table}[t]
  \centering
  \caption{Top-32 compression on ten training papers. All rows use the same
  frozen candidate--issue matrix. Oracle selectors access that matrix and are
  upper bounds, not deployable systems.}
  \label{tab:clustering-selection}
  \scriptsize
  \begin{tabular}{@{}lrrrr@{}}
    \toprule
    Selector & Strict & Weighted & Major & Precision \\
    \midrule
    Qwen pointwise, one/cluster & .350 & .402 & .643 & .191 \\
    Kimi tournament, one/cluster & \textbf{.390} & \textbf{.442} & \textbf{.727} & \textbf{.206} \\
    Cluster Oracle (hidden best) & .699 & .755 & .873 & .384 \\
    Unrestricted Top-32 Oracle & .739 & .793 & .940 & .406 \\
    Full 256 pool & .739 & .793 & .940 & .051 \\
    \bottomrule
  \end{tabular}
\end{table}

\subsubsection{Historical matching is not the goal}

Agreement with human reviews measures anticipation of a finite panel, not the
full value of AI feedback. Unmatched candidates include errors, but they can
also identify concrete issues the panel omitted. For \emph{OmniSpatial}, one
candidate flags different answer-extraction methods for reasoning and
non-reasoning models as a possible confound. For \emph{Regularization can make
diffusion models more efficient}, another asks how an asymptotic guarantee
supports the motivating few-step regime without a finite-$T$ threshold. Both
are specific and paper-grounded, yet neither crosses the historical-match
threshold. We therefore evaluate historical agreement separately from
independent validity; expert adjudication of the latter remains incomplete.

\section{Ablation study}
\label{sec:ablations}

We vary exploration, clustering, representative selection, and matching to
locate the main bottlenecks.

\subsection{Exploration policy}

Fine-grained persona quotas increase duplicate rejection from 37.1\% to 74.5\%,
while rigid section--category--persona slots return \texttt{no\_issue} 56.7\%
of the time. Together with adaptive stopping, these results show that local
novelty is not enough; exploration must value marginal pool coverage.

\subsection{Partitioning and representative selection}

Across eight clustering families, the one-per-cluster Oracle retains .755
weighted coverage versus .793 for the unrestricted Oracle. Of 160
clusters, 70 have no historical edge and 41 link to multiple issues, indicating
both splitting and merging errors, but only modest aggregate partition loss.

\begin{figure}[t]
  \centering
  \includegraphics[width=\linewidth]{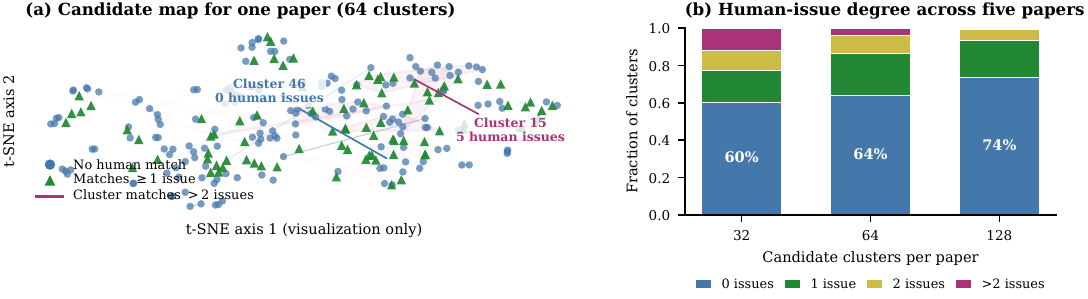}
  \caption{Candidate clusters and their historical-review matches.}
  \label{fig:cluster-diagnostics}
\end{figure}

Representative selection loses much more. With a fixed Ward partition, Qwen
pointwise selection and the best LLM tournament reach .402 and .442 weighted
coverage, versus .755 for the cluster Oracle. On five confirmation papers,
partitioning costs 6.1 points and Qwen selection costs 36.1. Only one candidate
captures the issue in 49.2\% of issue-bearing clusters, and Qwen score has .158
Spearman correlation with hidden utility. Retaining four candidates per
cluster raises the hidden ceiling to .709, locating the bottleneck in final
selection rather than shortlist construction.

\subsection{Matcher and target sensitivity}

At the same .70 threshold, Sol High marks 660 of 45,568 candidate--issue pairs
positive; DeepSeek V4 Flash marks eight ($\kappa=.015$). This scale collapse is
specific to the tested deployments but shows that valid outputs do not ensure
calibrated probabilities. Review atomization adds further noise by dropping
entities, qualifiers, or causal context, so measured misses combine pipeline
errors with matcher and target-extraction errors.

\subsection{Generator and judge interaction}

With provider labels removed, a Sol selector still chooses 319/320 Sol
candidates from mixed pools. Because Sol also supplies the evaluation matrix,
style recognition or same-family preference can confound ranking gains
\citep{panickssery2024selfrecognition}. Judge identity, prompt, threshold, and
generator-conditional error should therefore accompany judge-guided results.

\section{Conclusion}

Broad LLM generation can anticipate much of a historical reviewer panel:
the deduplication-and-refill pipeline reaches 78.7\% strict and 84.9\% weighted coverage in
our ten-paper diagnostic. The AI pool also contains concerns
outside the recorded reviews. This is the promise of pre-submission review:
useful feedback arrives while authors can still act, and need not be limited to
imitating one finite panel.

The open problem is trustworthy compression. Thirty-two concerns are enough
under a hidden oracle, and semantic clustering is useful, but present
representative selectors lose roughly 35 weighted-coverage points. Matcher and
atomization noise further obscure the true overlap. The right evaluation must
therefore report generation breadth, Top-$k$ loss, cost, judge robustness, and
two distinct outcomes: anticipation of human feedback and independent validity
of AI-only feedback. Under those constraints, LLM review can complement formal
peer review without replacing it.

\bibliographystyle{plainnat}
\bibliography{references}

\appendix
\section{Additional experimental details}

All evidence required for the main claims appears in the body. Frozen
machine-readable tables retain per-paper values, model and prompt versions,
candidate IDs, provider usage, and artifact hashes.

\section{Judge and annotation disclosure}

The headline matcher is a frozen GPT-5.6 Sol High prompt at temperature zero
and threshold .70. Raw probabilities, pair texts, structured outputs, and
cache keys are retained. A second DeepSeek prompt produced the scale-collapse
diagnostic and was never substituted into headline metrics. The blinded manual
Top-32 interface hides historical reviews and stores rankings separately by
annotator and paper. The pilot is not complete enough to estimate validity or
inter-annotator agreement; its only use here is to motivate a preregistered,
independent expert audit.

\section{Candidate-generation experiment inventory}

Table~\ref{tab:generation} records the broader sequence of generation
experiments behind the focused strategy comparison in the main paper. The
machine-readable values and their artifact provenance are stored in
\texttt{data/candidate\_generation\_metrics.json}; the publication figure is
regenerated with \texttt{scripts/plot\_candidate\_generation.py}.

\begin{table}[t]
  \caption{Candidate-generation experiments. All recall values use frozen
  atomic human issues and a same-issue threshold of $0.70$ unless noted.}
  \label{tab:generation}
  \centering
  \small
  \resizebox{\linewidth}{!}{%
  \begin{tabular}{p{2.2cm}p{3.1cm}p{7.4cm}}
    \toprule
    Experiment & Intervention & One-line result \\
    \midrule
    Persona discovery & Eight review-focus archetypes & Supplies structured
      exploration directions; downstream coverage benefit remains unproven. \\
    Sol vs. DeepSeek & 32 candidates/model, 93 papers & Individual recall is
      0.387/0.423; union recall is 0.535, showing complementary misses. \\
    Single vs. mixed 512 & DeepSeek-only vs. Sol+DeepSeek & Mixed recall is
      0.764 vs. 0.449; just 12.8\% of the single-model pool is unique. \\
    Persona submodes & Fixed persona--submode quotas & Better early prefixes,
      but 74.5\% duplicate rejection and zero completed 512 pools. \\
    Novelty stopping & Stop after two low-yield waves & 87.6\% fewer calls,
      but macro recall drops from 0.798 to 0.610. \\
    Structured exploration & Section--category--persona slots & More clean
      unique concerns, but 56.7\% \texttt{no\_issue} outputs. \\
    Eligibility adaptation & Retire repeatedly empty cells & Reallocation
      succeeds mechanically but generates near-duplicates in productive cells. \\
    \bottomrule
  \end{tabular}}
\end{table}

\begin{table}[t]
  \caption{Cost--coverage consequence of novelty-based early stopping. Counts
  and tokens are totals over the same ten-paper development cohort. Wall time
  was directly observed only for the adaptive run; n/r means not recorded.}
  \label{tab:adaptive-stopping}
  \centering
  \scriptsize
  \setlength{\tabcolsep}{4.2pt}
  \resizebox{\linewidth}{!}{%
  \begin{tabular}{lrrrrrr}
    \toprule
    Strategy & Retained & Requests & Tokens & Wall & Macro recall & Weighted \\
    \midrule
    Full dedup + refill & 5,120 & 1,905 & 96.61M & n/r & 0.798 & 0.849 \\
    Adaptive early stop & 1,125 & 236 & 7.45M & 15.0 min & 0.610 & 0.660 \\
    Relative change & $-78.0\%$ & $-87.6\%$ & $-92.3\%$ & --- & $-0.188$ & $-0.189$ \\
    \bottomrule
  \end{tabular}}
\end{table}

\section{Clustering artifact provenance}

The compact cluster graph and the two illustrative unmatched concerns are in
\texttt{data/clustering\_analysis.json}. They are exported from the frozen UI
bundle with \texttt{scripts/export\_clustering\_analysis.py}; the publication
figure is regenerated with \texttt{scripts/plot\_clustering\_analysis.py}.

\section{Secondary acceptance-judging diagnostic}

Outside the main review-generation story, we compared paper-only and sanitized
paper-plus-review outcome judgments on 15 training papers (10 Accept, five
Reject). Sol and Kimi improved with review prose, while DeepSeek and Inkling did
not improve hard accuracy. Because the cohort is small, schemas differ between
conditions, and current review notes may contain post-rebuttal edits, this is an
exploratory context-sensitivity result rather than an acceptance predictor.

\begin{table}[t]
  \caption{Binary Accept/Reject prediction on 15 training papers. ``A/R rec.''
  gives Accept and Reject recall. Reviews contain prose only.}
  \label{tab:acceptance-judging}
  \centering
  \scriptsize
  \setlength{\tabcolsep}{3.3pt}
  \begin{tabular}{@{}lrrrcrrr@{}}
    \toprule
    & \multicolumn{3}{c}{Paper only} && \multicolumn{3}{c}{Paper + reviews} \\
    \cmidrule(lr){2-4}\cmidrule(l){6-8}
    Judge & A/R rec. & Acc. & AUROC && A/R rec. & Acc. & AUROC \\
    \midrule
    DeepSeek V4 Flash & .30/.80 & .467 & .530 && .10/1.00 & .400 & .500 \\
    Kimi-K3 & .30/.40 & .333 & .460 && .70/.80 & .733 & .820 \\
    Inkling NVFP4 & .70/.60 & \textbf{.667} & \textbf{.640} && .30/.80 & .467 & .700 \\
    GPT-5.6 Sol High & .50/.40 & .467 & .520 && .70/1.00 & \textbf{.800} & \textbf{.920} \\
    \bottomrule
  \end{tabular}
\end{table}

\end{document}